\documentclass[a4paper, 10pt, conference]{IEEEtran}      
\IEEEoverridecommandlockouts
\usepackage[left=20mm,
        right=13.4mm,
        top=37mm,
        bottom=22mm]{geometry}
\usepackage{cite}
\usepackage{nth}
\usepackage{amsmath,amssymb,amsfonts}
\usepackage{algorithmic}
\usepackage{graphicx}
\usepackage{textcomp}
\usepackage{xcolor}
\usepackage{listings}
\usepackage{makecell}
\usepackage{ifthen}
\usepackage{amsmath}
\usepackage{amssymb}
\usepackage{amsfonts}
\usepackage{textcomp}
\usepackage{gensymb}
\usepackage{dashrule}
\usepackage{relsize}
\usepackage{comment}
\usepackage{textgreek}
\usepackage{accents}
\let\labelindent\relax
\usepackage{enumitem}
\usepackage{listings}
\usepackage{fancyhdr}
\usepackage{mathrsfs}
\usepackage{mathtools}
\usepackage{xfrac}

\usepackage{bm}
\usepackage{mathrsfs}
\usepackage{mathtools}
\usepackage{roundbox}
\usepackage{booktabs}
\usepackage[font=small]{caption}
\usepackage{subcaption}
\usepackage{hyperref}
\hypersetup{
    colorlinks   = true, 
    urlcolor     = [rgb]{0,0.40,0.85}, 
    linkcolor    = [rgb]{0,0.40,0.85}, 
    citecolor    = [rgb]{0,0.40,0.85}   
 }

{
\title{\vspace{5mm}Soft Fluidic Closed-Loop Controller for Untethered Underwater Gliders

\thanks{\hspace{-3mm}\textsuperscript{1} Department of Robotics Engineering, Worcester Polytechnic Institute, 100 Institute Road,  Worcester,  MA  01609,  USA.\\ \textsuperscript{2} Department of Mechanical Engineering, Worcester Polytechnic Institute, 100 Institute Road,  Worcester,  MA  01609,  USA.\\ \textsuperscript{3} Department of Electrical Engineering, Worcester Polytechnic Institute, 100 Institute Road,  Worcester,  MA  01609,  USA. \vspace{1mm}\\ \dag{} Indicates equal contribution \\ * To whom correspondence may be addressed: \href{mailto:mnemitz@wpi.edu}{mnemitz@wpi.edu}}}

\author{ Kalina Bonofiglio,\textsuperscript{1,\dag{}} Lauryn Whiteside,\textsuperscript{1,2,\dag{}} Maya Angeles,\textsuperscript{1 \dag{}} Matthew Haahr,\textsuperscript{1} Brandon Simpson,\textsuperscript{1} \\Josh Palmer,\textsuperscript{2} Yijia Wu,\textsuperscript{1} and Markus P. Nemitz\textsuperscript{1,2,3,*}}
}

\begin{document}

\maketitle

\bstctlcite{IEEEexample:BSTcontrol}
\thispagestyle{fancy}
\pagestyle{fancy}

\begin{abstract}
Soft underwater robots typically explore bio-inspired designs at the expense of power efficiency when compared to traditional underwater robots, which limits their practical use in real-world applications. We leverage a fluidic closed-loop controller to actuate a passive underwater glider. A soft hydrostatic pressure sensor is configured as a bang-bang controller actuating a swim bladder made from silicone balloons. Our underwater glider oscillates between the water surface and 4 m depth while traveling 15 m translationally. The fluidic underwater glider demonstrates a power efficiency of 28 $\sfrac{\text{mW}}{\text{m}}$. This work demonstrates a low-cost and power-efficient underwater glider and non-electronic controller. Due to its simple design, low cost, and ease of fabrication using FDM printing and soft lithography, it serves as a starting point for the exploration of non-electronic underwater soft robots.\smallskip
\end{abstract}


\section{Introduction}

\subsection{Traditional Underwater Gliders}
Over the last several decades, underwater gliders have gained popularity among autonomous underwater vehicles (AUVs) \cite{wood_autonomous_2009, bachmayer_underwater_2004}. Compared to other AUVs, underwater gliders can achieve greater traveling distances, lower power consumption, and improved cost effectiveness. Instead of using thrust propulsion with a propeller, underwater gliders use a change in buoyancy to travel large distances, oscillating between depths. Their improved power efficiency has made gliders promising technologies for tasks that involve underwater data collection.

\begin{figure}[hbt!]
    \centering
    \includegraphics[width=0.4\textwidth]{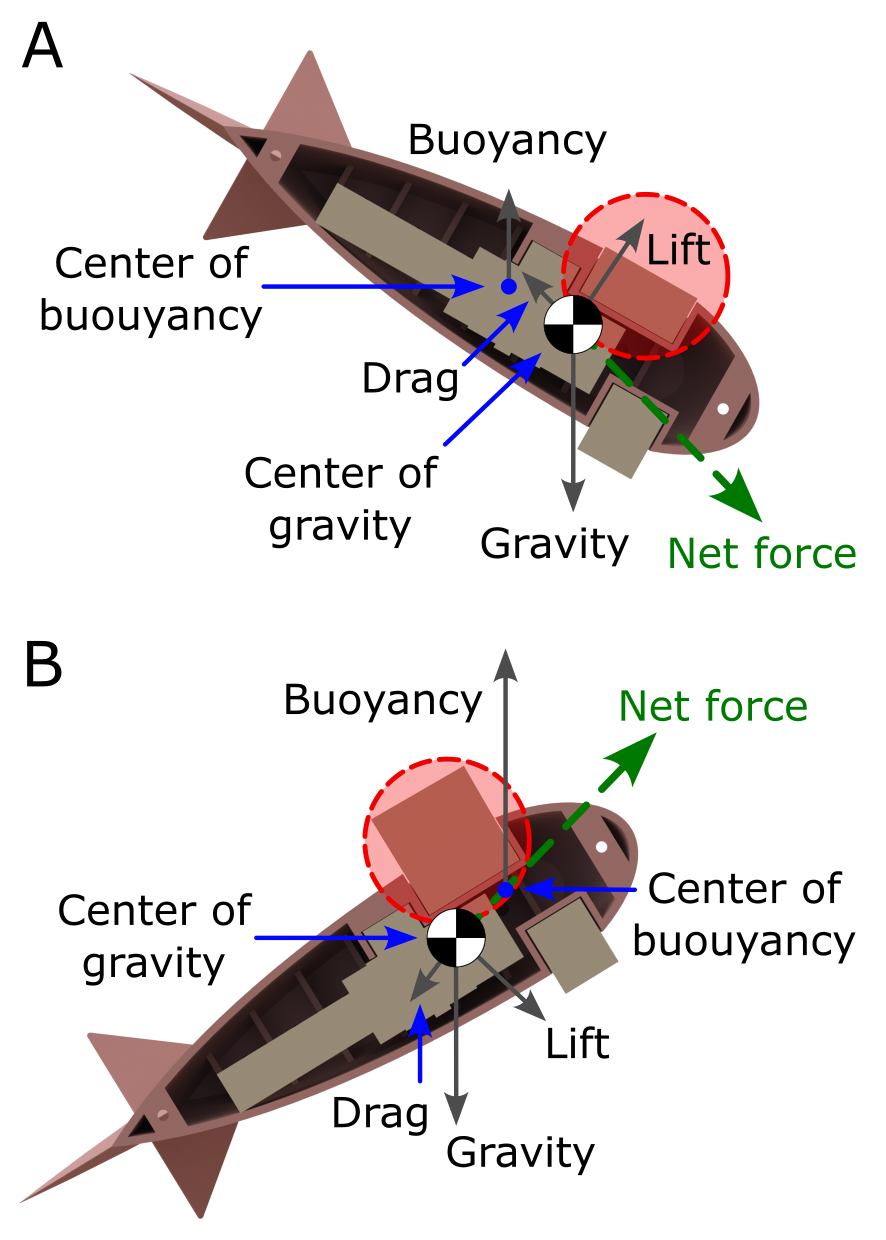}
    \caption{\textbf{Free Body Diagram of our Glider.} A) The swim bladder is deflated, lowering the buoyancy of the glider and shifting the center of buoyancy rearwards. The glider dives and the wing produces lift with a horizontal component causing the glider to translate forward.\hspace{3mm}B) The swim bladder is inflated, increasing the buoyancy of the glider and shifting the center of buoyancy forwards. The glider rises and the wing produces lift with a horizontal component causing the glider to translate forward.}
    \label{fig:FBD}
\end{figure}

Underwater gliders, a sub-category of AUVs, can be controlled either: (i) electrically and mechanically, (ii) using hybrid gliding and thrust propulsion, (iii) using thermal gradients, or (iv) leveraging wave and solar energy \cite{wood_autonomous_2009}. \textit{Slocum} and \textit{Seaglider} are well established underwater gliders and have become the industry standard in underwater gliding for control types i-iii \cite{eriksen_seaglider_2001, slocum_2017, webb_slocum_2001}. The \textit{Wave Glider} uses wave and solar energy as a renewable source of power (control type iv) \cite{manley_wave_2010, hine_wave_2009}.

\subsection{Untethered Underwater Soft Robots }
Soft underwater robots are cost effective options for ocean monitoring that can complement existing underwater systems \cite{aracri_soft_2021}. The inherent flexibility of soft materials used in soft robots makes them suitable for interactions with humans and delicate marine environments \cite{aracri_soft_2021}. The unique properties of soft materials has led to the development of bio-inspired actuators using actuation strategies including stimuli responsive materials (SRMs), chemical reactions, and fluidic actuation \cite{el-atab_soft_2020}. 

SRMs change shape or mechanical properties in the presence of specific stimuli \cite{bril_stimuli-responsive_2022}. Dielectric elastomer actuators (DEAs) and ionically conductive hydrogels are electrically responsive materials and have been implemented in soft underwater robots, including a deep sea soft underwater glider \cite{li_self-powered_2021} and an untethered fully soft fish \cite{li_fast-moving_2017, el-atab_soft_2020}. Thermally responsive materials have been successfully implemented in untethered robots; the bioinspired intermittent glider implements thermo-electric and pneumatic actuators for local buoyancy control \cite{lee_bioinspired_2022}. Thermally responsive materials have slow response times \cite{el-atab_soft_2020}. Combustion based soft actuators use explosive chemical reactions to produce high forces \cite{el-atab_soft_2020}, allowing for controlled and repeatable jet propulsion underwater \cite{keithly_cephalopod-inspired_2018}. While combustion actuators can achieve high forces, they demonstrate low actuation frequencies and a limited operational lifetime \cite{el-atab_soft_2020}. Fluidic based actuators have been implemented in bioinspired hybrid robots including the Robot Tuna \cite{anderson_maneuvering_2002} and SoFi \cite{katzschmann_hydraulic_2016, el-atab_soft_2020}. Hybrid refers to robots that combine soft and rigid materials. Hydraulic pumps move fluids between two-chamber systems, creating undulatory motions in the soft tailfin.

Power consumption and power efficiency are important in both, soft and traditional (rigid) untethered underwater robots \cite{aracri_soft_2021}. In this work, we define power efficiency as power consumption per distance traveled. Gliders have a higher power-efficiency when compared to other types of AUVs because they use discrete changes in buoyancy and fin pitch for propulsion instead of continuous thrust. We summarize the estimated power efficiencies of current underwater robots with different propulsion types including this work in \textbf{Table \ref{tab1}}.

\begin{table*}[t]
\begin{center}
\def\arraystretch{1.5}
\begin{tabular}{|c|c|c|c|c|c|}
\hline
System Name & \makecell{Propulsion\\Technology} & \makecell{Power\\Efficiency $\left[\sfrac{\text{mW}}{\text{m}}\right]$} & Gliding Range [m] & Gliding Depth [m] & Deployment Time [h] \\ \hline\hline
Seaglider \cite{queste_deployments_2012} & Mechanical / Electrical & $ 3.84 * 10^{-4\:\maltese}$  & $ 2.826 * 10^{6\:\maltese}$ & 1019 & 3144 \\ \hline
Slocum \cite{cooney_expanding_2016} & Hybrid gliding propulsion & $42.8125 \left[\sfrac{\text{J}}{\text{m}}\right] ^{\bigstar} $  & n/a & 100 & n/a \\ \hline
Tianjin University \cite{yang_thermal_2016}  & Thermal & $2.0697^{\,\maltese}$ & $6.77*10^5$ & 1000 & 648 \\ \hline
Wave Glider \cite{hine_wave_2009} & Wave and Solar & $ 2.16 * 10^{-2\:\maltese}$ & $ 3.982 * 10^{6\:\maltese}$ & n/a & 5928  \\ \hline
\makecell{Fast Moving\\Manta Ray \cite{li_fast-moving_2017}}  & \makecell{Ionic Hydrogel\\and DEA} & $42.10^{\,\maltese}$ & $128.7^{\,\maltese}$ & unknown & 3.25  \\ \hline
Wireless Flatfish \cite{lee_bioinspired_2022} & \makecell{Thermoelectric\\Pneumatic\\Actuator} & $3.236 * 10^{5\:\maltese}$  & $72$ & $10.5^{\,\maltese}$ & 1  \\ \hline
SoFi \cite{katzschmann_exploration_2018} & Hydraulic Pump & $178.67^{\,\maltese}$ & $296.8 \pm 5.1$ & 8.1 &  0.66 \\ \hline
\textbf{Our Implementation} & Fluidic circuit & 28 & 150 & 4 & 0.25  \\ \hline
\end{tabular}
\caption{\textbf{Comparison of Existing Systems} \\ \maltese{} indicates estimated values that are not directly provided in the cited paper\\ $\bigstar$ indicates that no time of deployment was available. Provided value is energy, not power.}
 
\label{tab1}
\end{center}
\end{table*}

\vspace{0.5mm}
\subsection{Fluidic Control Circuits}

Soft control circuits have been used as a substitute for electronic control circuits because of their light weight \cite{nemitz_soft_2020}, resistance to impact \cite{nemitz_soft_2020}, resistance to harsh environments \cite{nemitz_soft_2020}, simplicity and low cost \cite{preston_soft_2019}, and their safe interaction with humans \cite{preston_soft_2019}. Existing soft controllers rely on conductive materials \cite{zhou_universal_2012, garrad_soft_2019, wang_highly_2017, teng_integrating_2018}, chemical reactions \cite{wehner_integrated_2016, shepherd_using_2013}, or material instabilities that result in buckling behaviors. 

The bistable valve is a soft robotic equivalent of the CMOS transistor, meaning it only requires power to switch between two states \cite{rothemund_soft_2018}. It contains four inputs, two outputs, and a soft, snap-through membrane that switches between the two outputs. Two inputs determine the pressure of chambers on either side of the membrane and therefore determine the state of the membrane; the other two inputs are signal lines. The membrane kinks one of the input pressure lines and allows the other pressure to pass through to the output. A bistable valve has been configured as a switch \cite{rothemund_soft_2018}, a fluidic logic gate and sensor \cite{preston_digital_2019}, a memory device \cite{nemitz_soft_2020}, and an oscillator \cite{preston_soft_2019}. 

\begin{figure}[t!]
    \centering
    \includegraphics[width=0.47\textwidth]{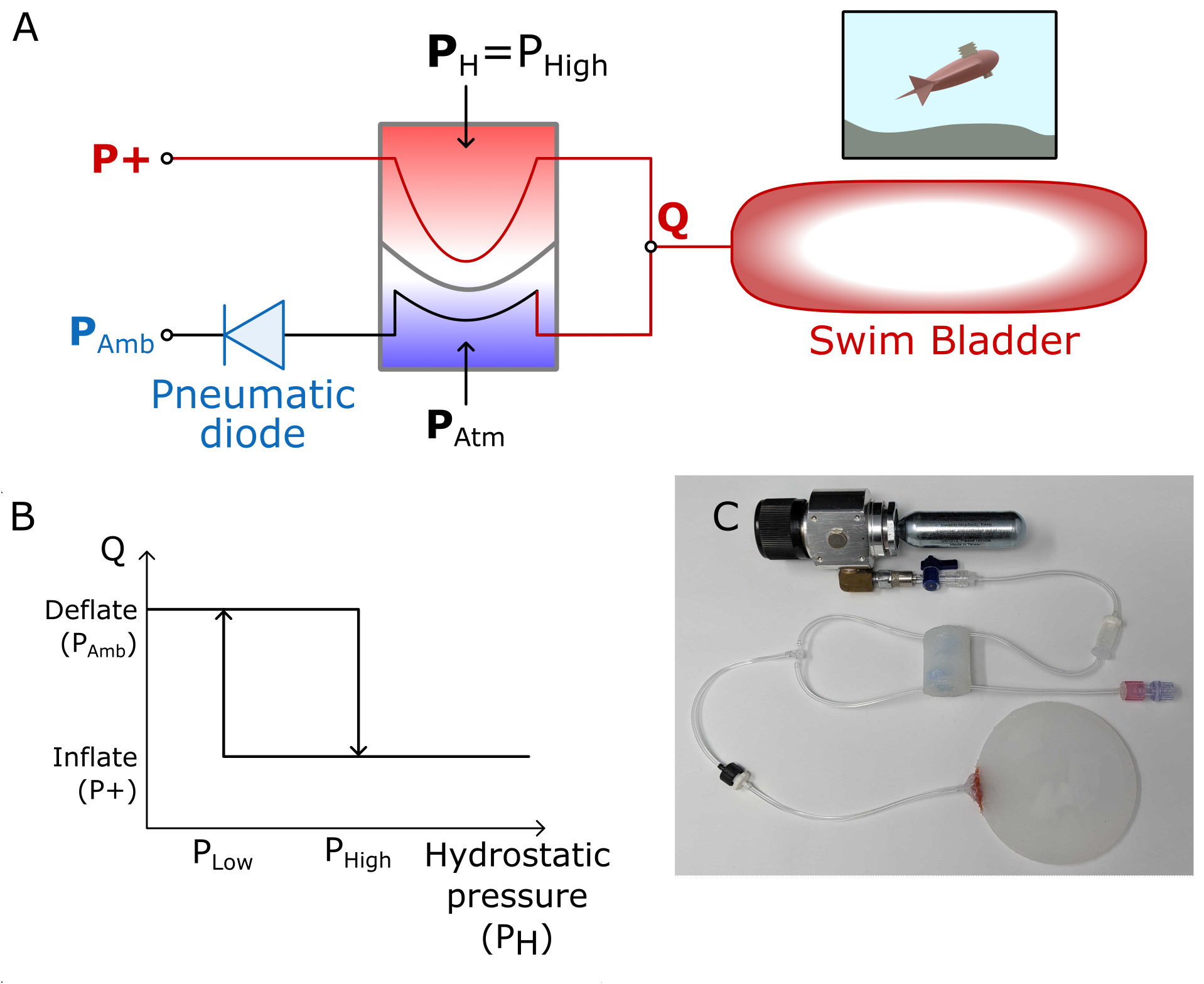}
    \caption{\textbf{Fluidic Controller.} A) Schematic of the fluidic circuit in the actuated state, where P\textsubscript{H} (the hydrostatic pressure) has exceeded the snap-through pressure of the membrane and the P+ supply pressure tube is unkinked. In this state, P+ supply pressure inflates the swim bladder.\hspace{3mm}B) Schematic of the bang-bang controller, which transitions from the deflated to the inflated state when the pressure exceeds the P\textsubscript{High} threshold and returns to deflation after crossing P\textsubscript{Low}.\hspace{3mm}C) Implementation of the fluidic circuit.}
    \label{fig:Circuit}
\end{figure}

\begin{figure*}[t!]
    \centering
    \includegraphics[width=0.9\textwidth]{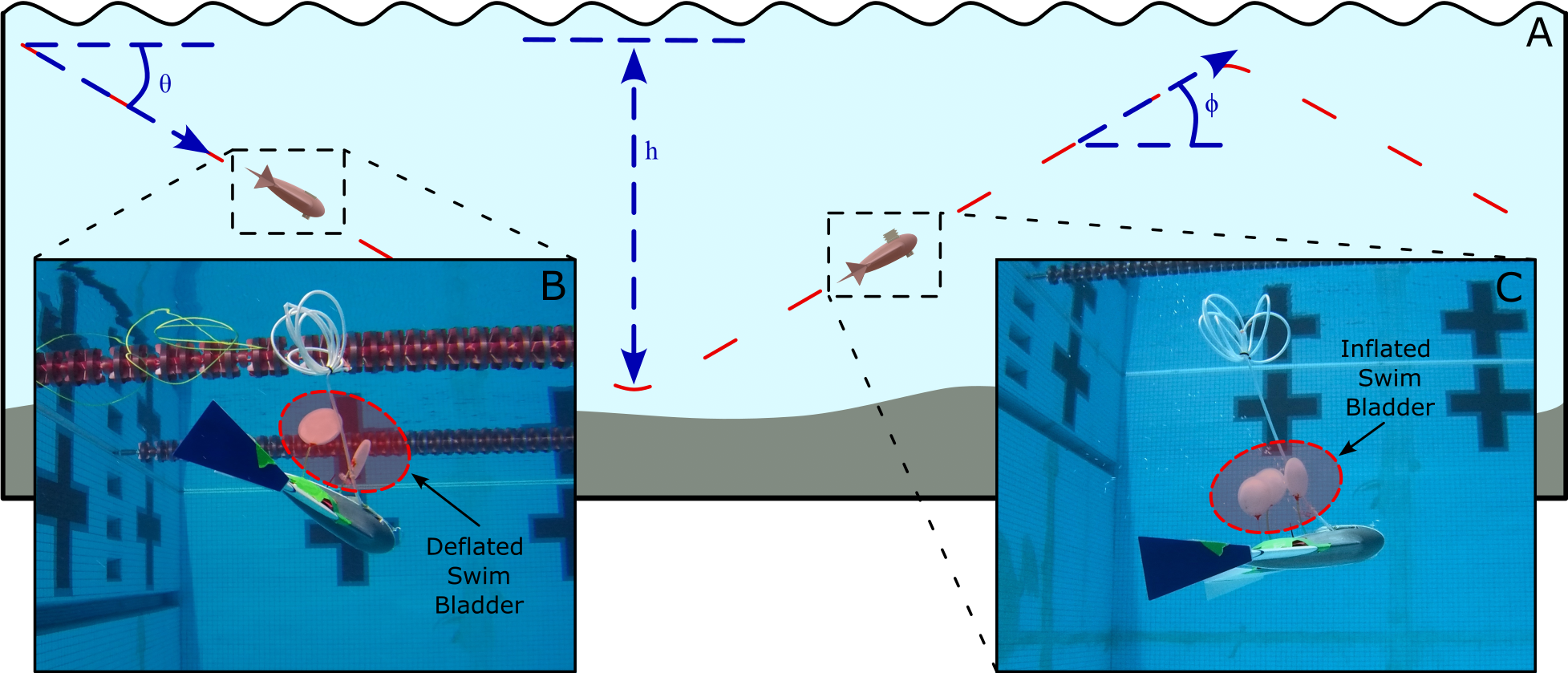}
    \caption{\textbf{Glider Overview.} A) The glider oscillates vertically in the water due to the change in buoyancy depending on the state of the swim bladder while the wings provide lift, causing horizontal motion. The glider descends at angle $\theta$ an ascends at $\phi$.\hspace{3mm}B) Glider diving due to the deflation of the swim bladder.\hspace{3mm}C) Glider rising due to the inflation of the swim bladder.}
    \label{fig:Iconic}
\end{figure*}

This work differs from previous implementations in that it combines the body design from traditional AUV gliders (\textbf{Figure \ref{fig:FBD}}) and a soft fluidic circuit (\textbf{Figure \ref{fig:Circuit}}). Leveraging the larger gliding distance and lower power requirements, we can implement an untethered, low cost, power efficient underwater glider (\textbf{Figure \ref{fig:Iconic}}). The contributions of this paper include: 

\begin{samepage}
    \begin{enumerate}[label = \arabic*)]
        \item Design and characterization of a soft bistable valve configured as a fluidic hydrostatic pressure sensor.
        \item Implementation of the hydrostatic pressure sensor into a fluidic circuit that controls the actuation of a swim bladder. 
        \item Implementation of the fluidic controller into a 3D printed \textit{blended-wing} inspired underwater glider. 
        \item Demonstration of an untethered underwater glider with an integrated fluidic controller that can perform 10 oscillations between water surface and a depth of 4 meters with a total range of 150 meters over 0.25 hours using a 16 gram CO\textsubscript{2} cartridge.   
    \end{enumerate}
\end{samepage}

\vspace{0.5mm}
\section{Design} <return>

\subsection{Blended Wing Glider Design}  

The blended-wing-body underwater glider (BWBUG) is an underwater glider design that blends the shape of the body into the wings to create a smooth transition between components; the design improves the hydrodynamic performance of AUV gliders. Sun et al. maximized the gliding distance of the BWBUG glider by evaluating ten shape parameters in simulation \cite{sun_shape_2017}. Out of these ten parameters, they found five key parameters that play a pivotal role in impacting gliding distance; these parameters include two relative span ratios, two relative thickness ratios, and a sweep back angle. They concluded a glider body with improved hydrodynamics comes at the cost of total internal volume, which is required for housing components.

\begin{figure}[ht!]
    \centering
    \includegraphics[width=0.45\textwidth]{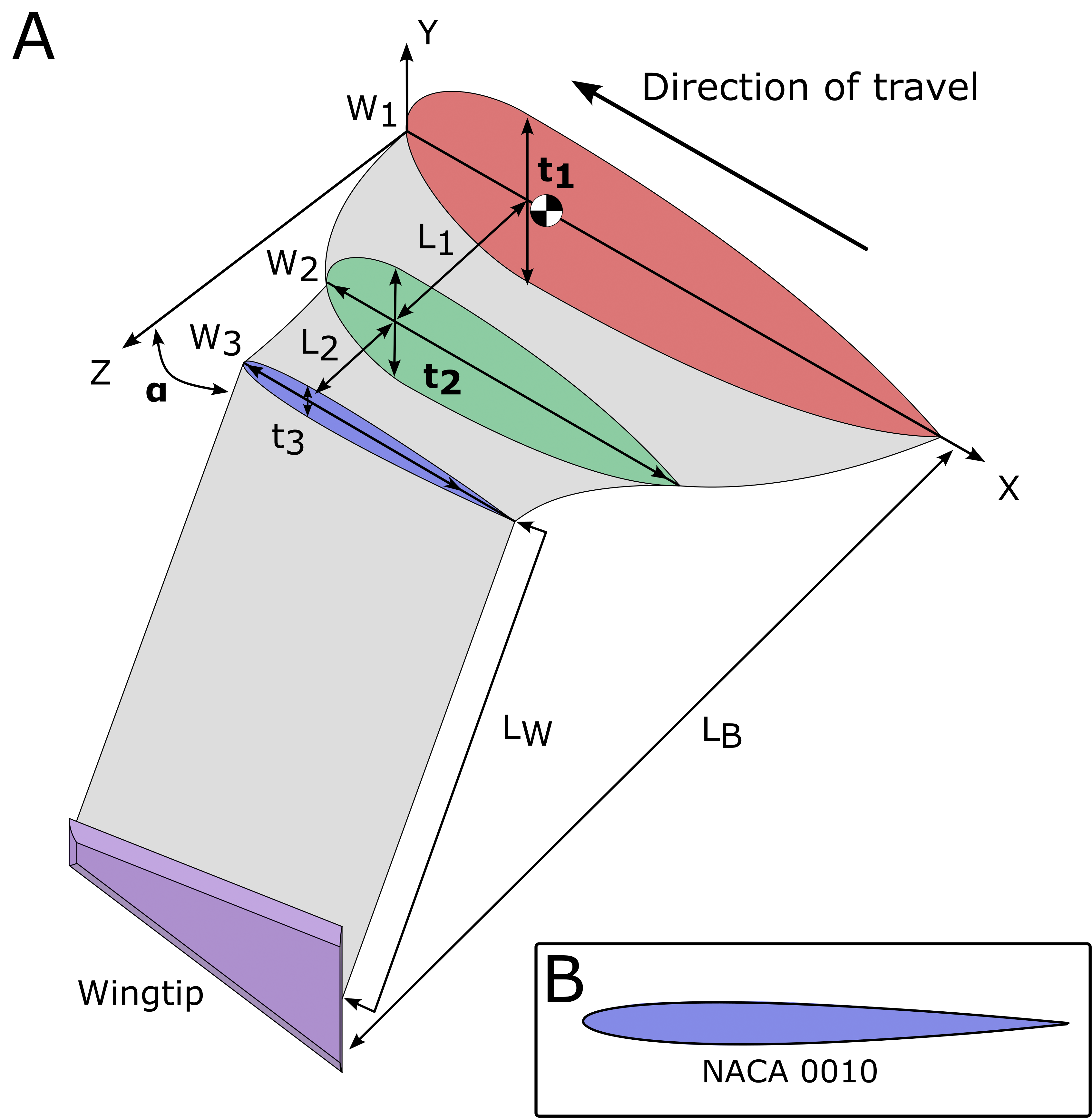}
    \caption{\textbf{Wing Diagram.} A) Surfaces t\textsubscript{1}, t\textsubscript{2}, t\textsubscript{3} indicate the midplane of the body, beginning of the wing root, and the end of the wing root and beginning of the wing respectively. All control components of the glider are between t\textsubscript{1} and t\textsubscript{2}. Angle α defines the sweep back angle of  the wings. Parameters L\textsubscript{1}, L\textsubscript{2} and L\textsubscript{B} defines the relative spanwise ratios for surfaces t\textsubscript{1}, t\textsubscript{2}, t\textsubscript{3}. B) Between surfaces t\textsubscript{3} and the wingtip, the cross section is constant, following the dimension of the NACA 0010 airfoil.}
    \label{fig:Wing}
\end{figure}

Our focus was to make the glider low cost, easy to assemble, and simple to manufacture. We derived aspects of our glider design from Sun et al.'s research findings to create greater hydrodynamic efficiency compared to traditional underwater gliders. We used the optimized sweepback angle from the BWBUG glider, and altered the measurement ratios including thickness and spanwise ratios. The thickness and spanwise ratios of our glider are different compared to the BWBUG glider due to our internal components and manufacturing capabilities. We added wingtips to stabilize the yaw motion of the system and improve linear motion \cite{lyu_winglet_2019}. \textbf{Figure \ref{fig:Wing}} illustrates the key parameters of our glider design.

\subsection{Fluidic Circuit} 


The closed loop fluidic circuit contains a pressure source (16g CO\textsubscript{2} cartridge), pressure regulator, bistable valve, swim bladder, and pneumatic diode (one-way valve) (\textbf{Figure \ref{fig:Circuit}}). We used a single bistable valve both as a hydrostatic pressure sensor and as a bang-bang controller. One chamber of the bistable valve was sealed with atmospheric pressure and the other was exposed to ambient hydrostatic pressure. The bistable valve has an internal hysteresis which is created by the difference between the initial \textit{snap-through} pressure of the membrane and the \textit{snap-back} pressure that is required to return to the original state. Preston et al. used this physical property to create an underwater profiler that oscillates between two depths \cite{preston_digital_2019}. We expanded on this application by co-developing a glider and fluidic circuit to achieve translational movement.

\begin{figure}[b]
    \centering
    \includegraphics[width=0.45\textwidth]{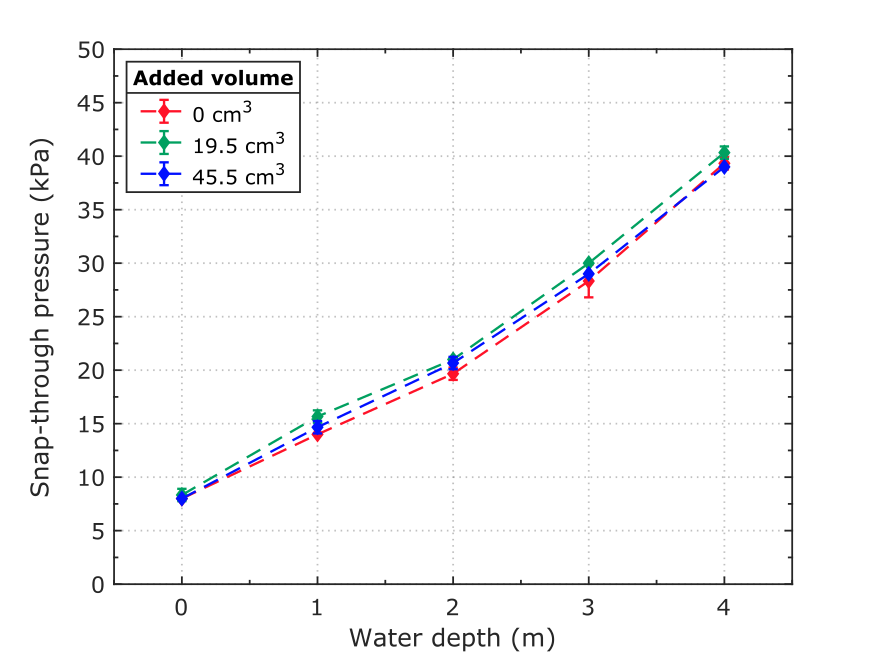}
    \caption{\textbf{Characterization of Bistable Valve as a Hydrostatic Sensor.} Evaluates the snap-through pressure for the bistable valve at 0-4m depth with three different sealed atmosphere chamber volumes created with a kinked length of tubing. Testing procedure: Seal the atmosphere tube at the desired additonal volume. Drop the system to specified depth. Add controlled pressure to the hydrostatic chamber until the membrane flips.} 
    \label{fig:characterization}
\end{figure}

The hysteresis properties of the soft bistable valve membrane are defined by membrane thickness and the opening angle of the membrane \cite{sun_shape_2017}. For our system, we used a 3mm thick membrane with an opening angle of $87.5^\degree$. This valve was predicted to snap-through at a pressure of 10 kPa and snap-back at a pressure of 1 kPa. We investigated the snap-through pressure of the valve at a range of depths with additional volumes attached to the sealed atmosphere chamber to determine the impact of hydrostatic pressure on membrane behavior (\textbf{Figure \ref{fig:characterization}}).

The hysteresis of the bistable membrane allows the bistable valve to be used as a bang-bang controller. The force to snap-through the membrane and unkink the supply pressure to inflate the swim bladder, is generated by the ambient hydrostatic pressure (\textbf{Figure \ref{fig:Circuit}}). The supply pressure inflates the swim bladder increasing the total buoyancy force experienced by the glider. When the glider reaches the hydrostatic pressure equivalent to the snap-back pressure of the membrane, the supply pressure is cut off, and the inflated swim bladders release pressurized air via a pneumatic diode. The glider then initiates its descent. The swim bladders consist of two layers of silicone separated by a thin sheet of wax paper. We fabricated the swim bladders from DragonSkin 10NV (SmoothOn Inc.) and created a waterproof seal between swim bladder and tubing for supply pressure using Sil-Poxy adhesive (SmoothOn Inc.). 


The power supply for our fluidic circuit is a 16g CO\textsubscript{2} cartridge connected to a pressure regulator. We set the output pressure of the pressure regulator to 40 kPa, which is the pressure that is required to inflate the swim bladder at a depth of 4 meters.

\section{Results}

\subsection{Description of the final system }
The net force on the glider determines whether it is neutrally, positively, or negatively buoyant. The net force can be described as the sum of gravitational force due to weight of the glider, buoyancy force due to the displacement of water by the glider, and buoyancy force dynamically changeable by an inflatable swim bladder (\textbf{Equation \ref{eq:Ftotal}}).

\begin{equation}\label{eq:Ftotal}
    F_{\text{net}} = (F_{B,\,\text{glider}} - F_{G,\,\text{glider}}) + F_{B,\,\text{swim bladder}}
\end{equation}

The gravitational and buoyancy forces of the glider are static during operation. They are controlled during the design phase and tuned experimentally such that the glider is just negatively buoyant (negative net force). The buoyancy force of the swim bladder is determined by the change in volume (\textbf{Equation \ref{eq:deltas}}).

\begin{equation}\label{eq:deltas}
    \Delta F_{B,\,\text{swim bladder}} = \Delta V \rho g
\end{equation}

It is important to consider depth $d$ at which the swim bladder inflates and adjust the supply pressure $P_{\text{supply}}$ for the inflation of the swim bladder. $P_{\text{swim bladder}}$ refers to the differential pressure that is required to inflate the swim bladder at atmospheric pressure (\textbf{Equation \ref{eq:pressures}}).

\begin{equation}\label{eq:pressures}
    P_{\text{supply}} = P_{\text{swim bladder}} + 10 \frac{\text{kPa}}{\text{m}} d
\end{equation}

The maximum buoyancy force that a swim bladder can create depends on the volume of water it can displace (\textbf{Equation \ref{eq:forces}}).

\begin{equation}\label{eq:forces}
    F_{\text{swim bladder}} = V_{\text{swim bladder}} \,\rho_{\text{water}}
\end{equation}

\textbf{Table \ref{tab:dimensions}} summarizes our design choices for the glider. 

\begin{table}[t]
    \centering
    \def\arraystretch{1.5}
    \begin{tabular}{|c|c|}
        \hline 
        Quantity & Value \\\hline \hline
        Total Glider Volume & $3861.12\;\text{cm}^3$ \\\hline
        Volume of Glider Cavity & $1696.91\;\text{cm}^3$ \\ \hline
        Swim Bladder Volume (per balloon) & $100\;\text{cm}^3$\\\hline
        Total Swim Bladder Volume & $300\;\text{cm}^3$\\\hline  
        Total Glider Weight & 3.722 kg\\\hline
    \end{tabular}
    \caption{\textbf{Glider Dimensions.}}
    \label{tab:dimensions}
\end{table}

\subsection{Power consumption}
Our glider system was able to dive to a depth of 4 meters and resurface while travelling 15 meters per cycle. The total cycle time was 90 seconds. The maximum travel range of our glider using a 16g CO\textsubscript{2} cartridge is 150 meters (10 cycles) under the assumption the valve dissipates air slowly and maintains a constant temperature (\textbf{Equation \ref{eq:gasLaw}}).

\begin{equation}\label{eq:gasLaw}
    d_{total} = \frac{P_{\text{cartridge}}}{P_{\text{swim bladder}} + 10 \frac{\text{kPa}}{\text{m}} d} \frac{V_{\text{cartridge}}}{V_{\text{swim bladder}}} d
\end{equation}







The total energy contained in one 16g CO\textsubscript{2} cartridge is 3.82 kJ \cite{wehner_pneumatic_2014}. Based on our system oscillation time and a total time traveled of 900 seconds and a distance of 150 meters, our power consumption is 4.2 W and our power efficiency is 28 $\frac{\text{mW}}{\text{m}}$.





\subsection{Cost}
We constructed our final system from low-cost materials using low-cost manufacturing methods. The final cost for one glider was under \$150 (\textbf{Table \ref{tab:cost}}). 

\begin{table}[h!]
    \centering
    \def\arraystretch{1.5}
    \begin{tabular}{|c|c|}
        \hline 
        Material or Component & Cost \\\hline \hline
        PLA & $\$27.86$ \\\hline
        DragonSkin 10 NV & $\$15.03$ \\ \hline
        Pressure Regulator & $\$79.99$\\\hline
        Tubing and Connectors & $\$10^{\,\maltese}$\\\hline  
        Weight & $\$8.20$\\\hline\hline
        \textbf{Total} & \textbf{$\$141.08$}\\\hline
    \end{tabular}
    \caption{\textbf{Glider Cost Breakdown.}\\ \maltese{} Indicates estimated cost}
    \label{tab:cost}
\end{table}

\section{Conclusion}
The power consumption and cost of underwater robots determine their real-world use. At this stage, soft robots have been demonstrated in real-world underwater environments, however, they lack the power efficiency of their rigid counterparts. The \textit{Seaglider} has the lowest power consumption in comparison to other rigid and soft gliders; however, it costs \$125,000, making it unsuitable for large-scale and distributed monitoring \cite{eriksen_seaglider_2001}.

In this work we present an underwater system that is both, low-cost and power efficient.  The proposed design costs under \$150 per unit due to the inexpensive 3D printing and soft-lithography fabrication techniques we used. Our glider design is a derivative of existing work and was augmented by implementing pitch and variable buoyancy control with a closed-loop fluidic bang-bang controller. The hydrodynamic performance and volume capacity of our glider could be further optimized to improve the power consumption of our system by, for example, reducing the overall weight of the glider, optimizing the change in volume of the swim bladder required for ascending and descending, or reducing the ascent and descent angles.

We conclude that our underwater robot is power efficient and low-cost. The fluidic, closed-loop control mechanism using a soft bistable valve configured as a hydrostatic pressure sensor is minimal but effective. Electronic systems achieving the same functionality are more expensive and require more complex fabrication processes including the watertight sealing of electronics. Moving forward, we hypothesize that fluidic controllers will be used in simple underwater systems and electronics will be added for sophisticated sensing and communication. 

\section*{Acknowledgment}

We thank the WPI Athletic and Recreation Center staff, the Worcester Academy, and the WPI Soft Robotics Lab for access to their facilities for testing.

\bibliographystyle{IEEEtran}
\bibliography{controls,references,extendedCite}


\end{document}